# The Best Templates Match Technique For Example Based Machine Translation

**T. El-Shishtawy (1) and A. El-Sammak (2)**
**Faculty of Engineering (Shoubra), Benha University, Cairo.**
**(1) shishtawy@hotmail.com    (2) sammaka@ipa.edu.sa**

**Abstract:**

It has been proved that large-scale realistic Knowledge Based Machine Translation (KBMT) applications require acquisition of huge knowledge about language and about the world. This knowledge is encoded in computational grammars, lexicons and domain models. Another approach – which avoids the need for collecting and analyzing massive knowledge- is the Example Based approach, which is the topic of this paper.

We show through the paper that using Example Based in its native form is not suitable for translating into Arabic. Therefore a modification to the basic approach is presented to improve the accuracy of the translation process. The basic idea of the new approach is to improve the technique by which template-based approaches select the appropriate templates. It relies on extracting, from a parallel Bilingual Corpus, all possible templates that could match parts of the source sentence. These templates are selected as suitable candidate chunks for the source sentence. The corresponding Arabic templates are also extracted and represented by a diredted graph. Each branch represents one possible string of templates candidate to represent the target sentence. The shortest continuous path or the most probable tree branch is selected to represent the target sentence. Finally the Arabic translation of the selected tree branch is generated.

**Keywords:** Arabic Machine Translation, Arabic Example Based Machine Translation, Parallel Bilingual Corpus, Morphology Analysis, Template Chunk, Correspondence Matrix, Directed Graph.

## أسلوب الأنماط الأكثر تطابقاً للترجمة الآلية المبنية على الأمثلة

لقد ثبت أن التطبيقات الحقيقية للترجمة الآلية المبنية على المعرفة تتطلب قدراَ هائلاً من المعرفة حول اللغة والنطاق الذي تطبق فيه. حيث يتم ترميز المعارف في صورة قواعد حسابية للنحو والصرف والمعاجم والقواميس. لذلك فإننا نقدم من خلال هذا البحث طريقة أخرى للترجمة باستخدام التطابق الأمثل للأنماط.

ويبين البحث أن استخدام أسلوب الترجمة بواسطة الأمثلة في صورته الأصلية لايصلح للترجمة إلى اللغة العربية. لذلك تم إقتراح بعض التعديلات للطريقة الأساسية لتحسين كفاءة عملية الترجمة. وتتلخص الطريقة الجديدة المقترحه في الاستعانة بمكنز لغوي ثنائي اللغة لاستخراج كل الأنماط المحتملة المطابقة لأجزاء من الجملة المراد ترجمتها بعد تقطيعها ، ويتم كذلك استخراج النمط العربي المقابل من خلال المكنز اللغوي ، ولتحسين الطريقة التي يتم بها إختيار الأنماط يتم تمثيلها بواسطة "المخطط الموجه" الذي يتكون من فروع تمثل كل واحدة منها أحد احتمالات أنماط الترجمة لجزء من الجملة. ثم يتم استنتاج نمط الجملة النهائية بالبحث عن أقصر (أو أفضل) مسار متصل خلال المخطط الموجه ، وفي النهاية يتم توليد الجملة المترجمة.

## 1. Introduction

Automatic translation between human languages (`Machine Translation') is a scientific dream of enormous social, political, and scientific importance. It was one of the earliest applications suggested for digital computers, but turning this dream into reality has turned out to be a much harder. And in spite of many problems, some degree of automatic translation is now available, and it is likely that during the next decade the bulk of routine technical and business translation will be done with some kind of automatic translation tools [1].

Unfortunately, Arabic language is not involved in MT progress. Very few systems do exist, but they lack an open architecture of the underlying theoretical basics of Arabic language. This lack - hidden, or missed - do not give a good picture of the state of MT that involves Arabic language.

It has been proved that large-scale realistic Knowledge Based Machine Translation (KBMT) applications require acquisition of huge knowledge about language and about the world. This knowledge is encoded in computational grammars, lexicons and domain models [2]. Some researchers seek ways of bringing down the price of knowledge acquisition by applying ways of automatically or semi-automatically extracting relevant information from machine-readable dictionaries [3] or text corpora [4]. Another approach – which avoids the need for collecting and analyzing massive knowledge- is the Example Based approach.

## 2. Example Based Machine Translation (EBMT)

Due to the increasing availability of large amounts of machine-readable textual material, a number of research groups investigate the possibilities for `empirical' approaches to machine translation.

Example-based translation is essentially translation by analogy. An Example-Based Machine Translation (EBMT) system is given a set of sentences in the source language (from which one is translating) and their corresponding translations in the target language, and uses those examples to translate other, similar source-language sentences into the target language [5]. The basic premise is that, if a previously translated sentence occurs again, the same translation is likely to be correct again [6]. It takes the advantage of aligned parallel corpora with a large number of short aligned text structures (Scania Corpora), to produce translation equivalent between English and Swedish.

In the `translation by analogy', or `example-based' approach, there are no mapping rules, only procedure which involves matching against stored example translations. The basic idea is to collect a bilingual corpus of translation pairs and then use a best match algorithm to find the closest example to the source phrase in question. This gives a translation template, which can then be filled in by word-for-word translation [7].

A main system which is based on EBMT is the Pangloss MT project [8] which translates from Spanish to English. The strategy is based on matching between chunks at word level. The chunks can be miss ordered or different in lengths which is not suitable for Arabic language.

Using EBMT in its native form, which is based on exact string matching, may include many complications when translating into Arabic. First, a huge parallel corpus is required to ensure matching of any input English sentence while translating at a sentence level. In this research, the translation is done at a smaller granularity level – which is called the chunk level. Second, mapping one-to-one to the sequence of words between the source language and Arabic as target language, will results in inaccurate translation. This is because the changes of the Arabic word (nouns, verbs, adjectives) according to the accompanied gender, count, and tense. Therefore, the aim of this paper is to seek for another solution which takes into account the features of Arabic language.

## 3. The Best-Template Match Approach

Another approach to example-based translations is to use a template-based model instead of string-based one. In this approach the system depends on template-based syntactic matching of the input English sentence, and the templates stored in a database. Once a matching occurs, the system gradually modifies the corresponding Arabic part to produce correct translation. It has the advantage that they require little examples in the bilingual parallel corpus archive to generate good translations. The template-based translation approach suffers from the problem of inaccurate or bad selection of templates for the source sentence. This affects the accuracy of the translation process. The proposed new approach is motivated to improve the translation accuracy by improving the technique by which template-based approaches select the appropriate templates. .

In the current work, we make use of a pre-built parallel bilingual corpus [9]. This corpus is organized as set of chunks in English language and their corresponding Arabic ones. Organizing the parallel corpus at chunk level, which are smaller building blocks of sentence, will be more flexible in translating different combinations of chunks using the same training corpus. Also, the proposed system relies on a rich on-line bi-directional dictionary, which was designed and prepared previously for ACLP-Machine Translation [10]. The dictionary contains about 40,000 English words and their corresponding translations into Arabic.

The basic idea of the new approach is to extract, from a parallel bilingual corpus, all possible templates that could match parts of the source sentence. These templates are selected as suitable candidate chunks for the source sentence. The corresponding Arabic templates are also extracted. To overcome the problem of inaccurate or bad selection of templates, the found chunks are represented in a tree structure. Each tree branch represents one possible string of templates candidate to represent the target sentence. The shortest continuous path is selected to represent the target sentence. Finally the Arabic translation of the selected tree branch is generated.

## 4. The Proposed Model

The algorithm of the proposed translation model is shown in figure (1). It starts with a morphological analysis (word category and lexical attributes) of the given English sentence. This valuable information will be used to extract all possible English chunk templates that could match possible tags of the given English sentence. The extracted templates are achieved by consulting the English-Arabic Parallel Corpus. Then the occurrence of each word of the given sentence within the candidate templates is tested. A correspondence matrix represents the result of this test is constructed. The corresponding matrix is tuned by removing useless templates and adding dummy templates. Set of template groups are extracted from the tuned correspondence matrix. Each group represents a possible path that covers all words of the given sentence. These paths are represented by a directed graph. The shortest path in the graph is selected as a string of chunk templates that best describes the given sentence. The corresponding string of Arabic chunk templates is extracted by consulting the English-Arabic corpus. Finally, the Arabic sentence of these chunks is generated, which represents the suggested translation of the given English sentence.

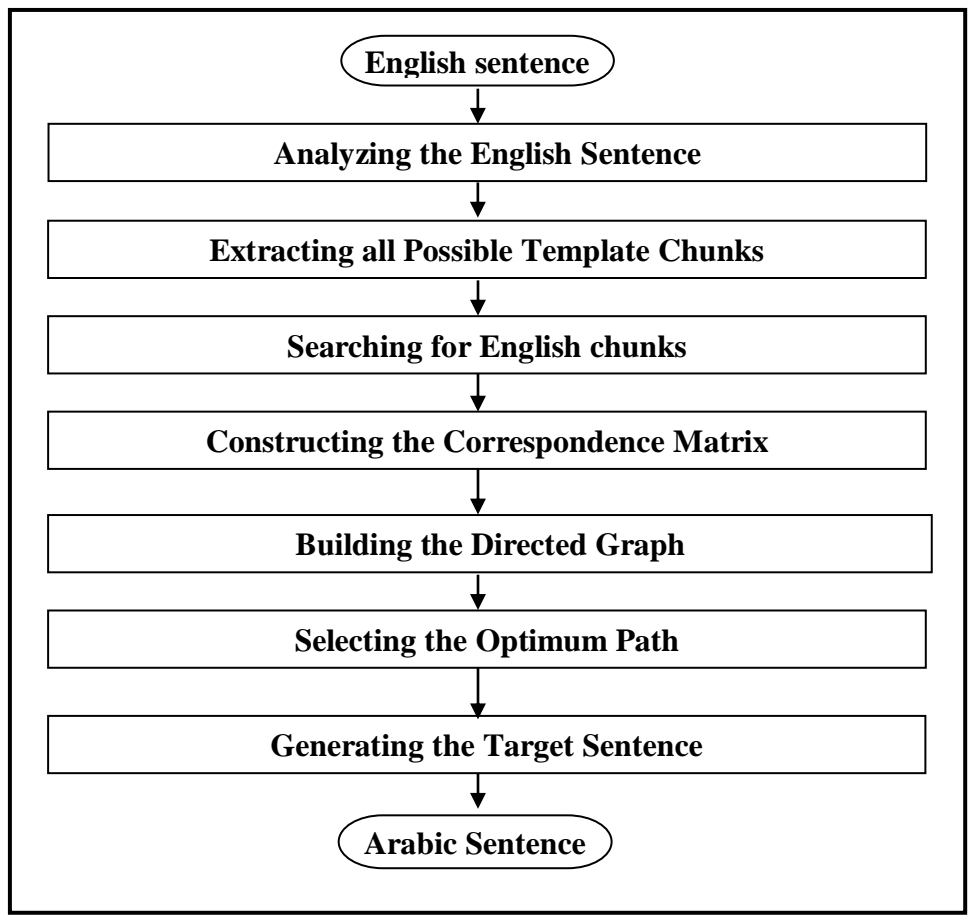

**Figure 1: The Best Template Match EBMT Algorithm**

Detailed description of the different phases of the proposed model is as follows:

*4.1 Analysis of Source Sentence*
    The given English sentence is analyzed by the following algorithm:
        The input sentence is segmented into word tags.
        For each word
            Search for the word in the English dictionary [10]
                If it is found
                      Extract the word lexical attributes, i.e.
                      (Category, tense, count, gender, …. )
                Else strip a prefix from the word (if found) and repeat
                Else strip a suffix from the word (if found) and repeat
        End For

*4.2 Extracting all Possible Template Chunks*
    The sequence of words (word category and its lexical attributes) constituting the input English sentence (w1 w2 w3 … wn) is applied to the following algorithm to extract all the possible template chunks:
      a. Start with the first word as one possible chunk.
      b. Search bilingual corpus for matching and extract its corresponding Arabic chunk
      c. Successively add next words –one by one- to the possible chunk, and repeat searching.(b) until the end of the sentence.
      d. Exclude the first word from the sentence and repeat (b) and (c)..
      e. Repeat (d) until the last word has been reached as the last chunk.

The output chunks consist of one or more consecutive words of the input English sentence and take the form:
$Ch\_tem_1 = \{w1\}$
$Ch\_tem_2 = \{w1, w2\}$
$Ch\_tem_3 = \{w1, w2, w3\}$
$Ch\_tem_n = \{w1, w2, w3,.....,wn\}$
$Ch\_tem_{n+1} = \{w2\}$
$Ch\_tem_{n+2} = \{w2, w3\}$
………..
$Ch\_tem_{2n-1} = \{w2, w3,.....,wn\}$
………..
$Ch\_tem_{n(n+1)/2} = \{wn\}$

*4.3 Searching for English chunks*
    In this phase we will search for all possible sentence chunk templates in the bilingual corpus. This can be achieved by matching the extracted English chunk templates with those stored in the bilingual parallel corpus. If the corpus template exactly matches the current sentence chunk template, the English chunk and its corresponding Arabic template are stored.

*4.4 Constructing the Correspondence Matrix*

The next step of our approach is to check the existence of the candidate English chunks within the English-Arabic parallel corpus. The result of the found chunks and their word coverage is then represented using a correspondence matrix. The correspondence matrix maps the coverage of the found English chunks, with the input sentence words. In other words, the contribution of each template chunk would be evaluated by describing the role of each template in representing part of the given sentence. This relation is best described by a correspondence matrix. The matrix rows represent the candidate chunks templates. The matrix columns represent the words patterns of the given sentence. The matrix cells take the values "1" or "0" according to the presence or absence of each word pattern within the found chunk template. The relation between chunks and words are many-to-many; i.e., chunk template can cover more than one word pattern. Also, the word pattern can belong to more than one chunk template.

Our goal is then to search through the correspondence matrix for the best continuous chunks that cover the whole input English sentences. Table (1) shows different cases for an English sentence. On analyzing the correspondence matrix, one might find five different types of chunks: repeated, unreachable, dead-end, dummy and normal chunks. These chunks are treated as follows:

1- Repeated chunks are removed except one. For example Ch_tem1, Ch_tem4, and Ch_tem7 are repeated chunks that cover the same sequence of words
2- Useless template chunks which do not affect the process of selecting the proper templates are deleted such as unreachable and dead-end chunks.
3- Also one might detect empty columns, which means that there is a discontinuity between chunks. Dummy chunks are inserted to overcome this discontinuity.

| Chunk Templates | Word Patterns | | | | | | | | | | |
|---|---|---|---|---|---|---|---|---|---|---|---|
| | wp1 | Wp2 | wp3 | wp4 | wp5 | wp6 | wp7 | wp8 | wp9 | wp10 | wp11 |
| Ch_tem 1 | ▓ | ▓ | | | | | | | | | |
| Ch_tem 2 | ▓ | ▓ | ▓ | ▓ | | | | | | | |
| Ch_tem 3 | | | | ▓ | ▓ | ▓ | ▓ | | | | |
| Ch_tem 4 | ▓ | ▓ | | | | | | | | | |
| Ch_tem 5 | | | | | | | | ▓ | ▓ | | |
| Ch_tem 6 | | | | | | ▓ | | | | | |
| Ch_tem 7 | ▓ | ▓ | | | | | | | | | |
| Ch_tem 8 | | | | | | | | | | ▓ | ▓ |
| Ch_tem 9 | | | | | | ▓ | ▓ | | | | |
| Ch_tem 10 | | | | | | | ▓ | ▓ | ▓ | | |
| Ch_tem 11 | | | | | | | | ▓ | ▓ | ▓ | ▓ |
| Ch_tem 12 | | | | | | | | | ▓ | ▓ | ▓ |
| Ch_tem 13 | | | | | ▓ | | | | | | |

**Table 1: Correspondence matrix**

The matrix shows several situations:

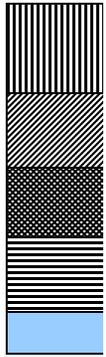

1. **Repeated chunks**: Chunks which cover the same number and sequence of words.
2. **Unreachable chunks**: chunks which can not be reached since it starts where no other chunks end.
3. **Dummy chunks:** Inserted chunks to overcome discontinuity
4. **Dead end chunks:** chunks when reached lead to a dead end, since there is no other chunks start at its end.
5. **Normal Chunks.**

### *4.5 Building the directed graph*

The best chunk's path can be detected by building a directed graph. In the graph, each node represents one word of the input sentence, and each branch represents a candidate chunk. The directed graph which represents the corresponding matrix is shown in Figure (2), where d1,d2,… define dummy chunks, 1,2,3,… represent words of the input sentence, and Ch_tem1, Ch_tem2,… are template chunks. The problem of complete translation is now reduced to finding an optimum path through the graph from first to final nodes (words). However in many real cases, there is discontinuity of the graph. For example, pronouns which are not found in parallel corpus can lead to such gap. To overcome this problem, we insert dummy chunks at all dead nodes, which are nodes without output branch (chunk). Dummy chunk connects dead nodes to their next neighborhood ones.

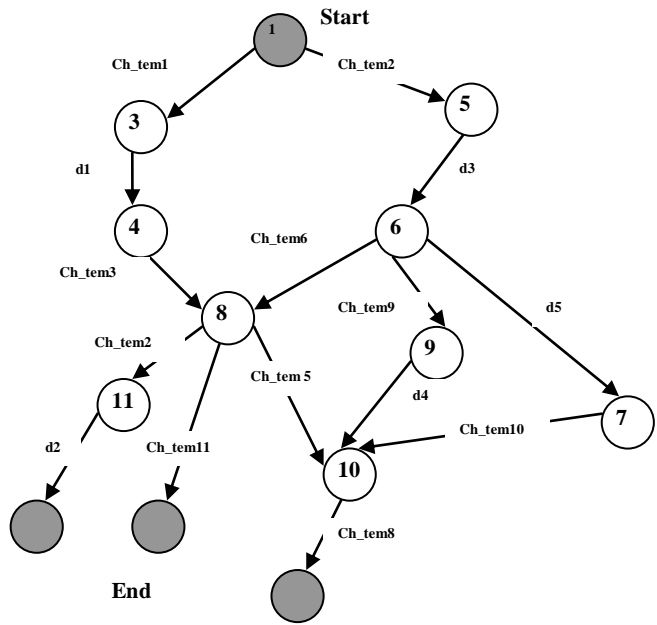

**Figure 2: Directed graph of Table 1**

## 4.6 Selecting the optimum Path

On examining directed graph, one might find that there is more than one possible path of chunk templates that can cover the whole word patterns of the given sentence. All the possible paths would be represented by an output table. Each row represents one possible path of chunk templates. Table (2) summarizes all possible paths for the previous graph.

The optimum path is selected as the shortest one with minimum dummy chunks. By examining different experimental cases, we found the following criteria:
- Longer chunks are preferred
- Paths with minimum number of dummies are preferred.

| Path No | Description | No. of Dummies | No. of Chunks |
|---|---|---|---|
| a) | Ch_tem1, d1,Ch_tem3,Ch_tem2,d2 | 2 | 3 |
| b) | Ch_tem1,d1,Ch_tem3,Ch_tem11 | 1 | 3 |
| c) | Ch_tem1,d1,Ch_tem3,Ch_tem5,Ch_tem8 | 1 | 4 |
| d) | Ch_tem2,d3,Ch_tem6,Ch_tem2,d2 | 2 | 3 |
| e) | Ch_tem2,d3,Ch_tem6,Ch_tem11 | 1 | 3 |
| f) | Ch_tem2,d3,Ch_tem6,Ch_tem5,Ch_tem8 | 1 | 4 |
| g) | Ch_tem2,d3,Ch_tem9,d4,Ch_tem8 | 2 | 3 |
| h) | Ch_tem2,d3,d5,Ch_tem10,Ch_tem8 | 2 | 3 |

**Table 2: all possible paths for Figure 2**

To implement these criteria, we use a simple heuristic that gives advantages to complete paths without dummy chunks. If there is more than one path with no dummies, we select the shortest-which contains minimum number of chunks. In cases, where there are dummies in all paths, we select the one with the smallest number of dummies.

In the given example, all paths contain dummy chunks. Selecting paths (b), (c), (e), and (f) as candidates containing only one dummy chunks. Among these paths, we select the paths (b) and (e) as optimum paths. This means that there are two possible translations for the given sentence.

## 4.7 Extracting the corresponding Arabic templates

The aim of this phase is to extract the Arabic templates corresponding to the English chunk templates of the selected path(s). This could be achieved by consulting the Bilingual Parallel Corpus. The extracted Arabic chunk templates are stored in the same sequence of the selected path. Table (3) shows the resultant Arabic chunk templates corresponding to English ones. During extraction, dummies are left unchanged in both sides. Actually, dummies represent unfound word(s) in parallel corpus, which may be due to pronouns or incomplete corpus. These unfound words are presented as English words inside the translated sentence.

|   | English Chunk Template Path | Corresponding Arabic |
|---|---|---|
| b) | Ch_tem1,d1,Ch_tem3,Ch_tem11 | ACh_tem1,d1,ACh_tem3,ACh_tem11 |
| e) | Ch_tem2,d3,Ch_tem6,Ch_tem11 | ACh_tem2,d3,ACh_tem6,ACh_tem11 |

**Table 3: The resultant Arabic chunk templates**

*4.8 Generating the Target Sentence*

A generation module is required to substitute Arabic templates with actual Arabic chunks. The inputs to this phase are the English chunks and their corresponding Arabic templates. Actually, Arabic chunks can't produced by word-to-word substitution of English chunk word's translation due to differences between English and Arabic surface structures. Therefore, Arabic template takes the form of command, which when applied to Arabic word, will produce the correct translation. Usually, the Arabic template is in the form of the following command:

"add [xx] , class[yy] , add[zz] "

Where xx & zz are optional and represent any added words or prefixes and suffixes part of the word yy, where yy is mandatory and describe the Arabic word category. The following algorithm is used to generate the Arabic chunks:

For each English chunk do
   If Chunk is dummy then
     Copy it to Arabic chunk
   Else
    For each English word do
      Get its translation according to corresponding accompanied Arabic attributes
        For each Arabic template do
          Execute the "add-class-add" command.

**5. Example 1**
Assume the English sentence to be translated is "*the Proteins are necessary for building our bodies*"

*5.1 Analyzing the English sentence*

| Word | Lexical Attribute |
|---|---|
| the | art |
| proteins | n [pl ,f] |
| are | be [p ,pl] |
| necessary | adj |
| for | prep |
| building | v [ing] |
| our | poss [pl ,m ,1] |
| bodies | n [pl ,f] |

*5.2 Extracting all Possible Template Chunks*

| All Possible English Template Chunks |
|---|
| art [def] |
| art [def]  n [pl ,f] |
| art [def]  n [pl ,f]  be [p ,pl] |
| art [def]  n [pl ,f]  be [p ,pl]  adj |
| art [def]  n [pl ,f]  be [p ,pl]  adj  prep |
| art [def]  n [pl ,f]  be [p ,pl]  adj  prep  v [ing] |
| art [def]  n [pl ,f]  be [p ,pl]  adj  prep  v [ing]  poss [pl ,m |
| art [def]  n [pl ,f]  be [p ,pl]  adj  prep  v [ing]  poss [pl ,m |
| n [pl ,f] |
| n [pl ,f]  be [p ,pl] |
| n [pl ,f]  be [p ,pl]  adj |
| n [pl ,f]  be [p ,pl]  adj  prep |
| n [pl ,f]  be [p ,pl]  adj  prep  v [ing] |
| n [pl ,f]  be [p ,pl]  adj  prep  v [ing]  poss [pl ,m ,1] |
| n [pl ,f]  be [p ,pl]  adj  prep  v [ing]  poss [pl ,m ,1]  n [pl |
| be [p ,pl] |
| be [p ,pl]  adj |
| be [p ,pl]  adj  prep |
| be [p ,pl]  adj  prep  v [ing] |
| be [p ,pl]  adj  prep  v [ing]  poss [pl ,m ,1] |
| be [p ,pl]  adj  prep  v [ing]  poss [pl ,m ,1]  n [pl ,f] |
| adj |
| adj  prep |
| adj  prep  v [ing] |
| adj  prep  v [ing]  poss [pl ,m ,1] |
| adj  prep  v [ing]  poss [pl ,m ,1]  n [pl ,f] |
| Prep |
| prep  v [ing] |
| prep  v [ing]  poss [pl ,m ,1] |
| prep  v [ing]  poss [pl ,m ,1]  n [pl ,f] |
| v [ing] |
| v [ing]  poss [pl ,m ,1] |
| v [ing]  poss [pl ,m ,1]  n [pl ,f] |
| poss [pl ,m ,1] |
| poss [pl ,m ,1]  n [pl ,f] |
| n [pl ,f] |

## 5.3 Searching for English chunks

| No | English chunk | English template |
|---|---|---|
| 1 | for getting | prep  v [ing] |
| 2 | for eating | prep  v [ing] |
| 3 | in playing | prep  v [ing] |
| 4 | girls | n [pl ,f] |
| 5 | the minerals | art [def]  n [pl ,f] |
| 6 | for feeding | prep  v [ing] |
| 7 | the proteins | art [def]  n [pl ,f] |
| 8 | necessary | adj |
| 9 | for building | prep  v [ing] |
| 10 | our bodies | poss [pl ,m ,1]  n [pl |
| 11 | the carbohydates | art [def]  n [pl ,f] |
| 12 | the fats | art [def]  n [pl ,f] |
| 13 | necessary | adj |
| 14 | the vitamins | art [def]  n [pl ,f] |

## 5.4 Constructing the Correspondence Matrix

| No | English template | w1 the art | w2 proteins n [pl ,f] | w3 are be [p | w4 necessary adj | w5 for prep | w6 building v [ing] | w7 our poss | w8 bodies n [pl |
|---|---|---|---|---|---|---|---|---|---|
| 1 | prep  v [ing] | | | | | ▓ | ▓ | | |
| 2 | prep  v [ing] | | | | | ▓ | ▓ | | |
| 3 | prep  v [ing] | | | | | ▓ | ▓ | | |
| 4 | n [pl ,f] | | ▓ | | | | | | |
| 5 | art [def]  n [pl ,f] | ▓ | ▓ | | | | | | |
| 6 | prep  v [ing] | | | | | ▓ | ▓ | | |
| 7 | art [def]  n [pl ,f] | ▓ | ▓ | | | | | | |
| 8 | adj | | | | ▓ | | | | |
| 9 | prep  v [ing] | | | | | ▓ | ▓ | | |
| 10 | poss [pl,m,1] n | | | | | | | ▓ | ▓ |
| 11 | art [def]  n [pl ,f] | ▓ | ▓ | | | | | | |
| 12 | art [def]  n [pl ,f] | ▓ | ▓ | | | | | | |
| 13 | adj | | | | ▓ | | | | |
| 14 | art [def]  n [pl ,f] | ▓ | ▓ | | | | | | |

Tuning the correspondence matrix is as follows:

The third word has no correspondence with any chunk, so dummy chunk (15) is added.
The chunks (2, 3, 6, and 9) are repeated with chunk (1), so they are deleted.
The chunks (14, 12, 11, and 7) are repeated with chunk (5), so they are deleted.
The chunk (13) is repeated with chunk (8), so it is deleted.
The chunk (4) is unreachable, so it is deleted.

The corresponding matrix after tuning is:

| N | English template | w1 | w2 | w3 | w4 | w5 | w6 | w7 | w8 |
|---|---|---|---|---|---|---|---|---|---|
|   |   | the | proteins | are | necessa | for | building | our | bodies |
|   |   | art | n [pl ,f] | be [p | adj | pre | v [ing] | poss [pl | n [pl |
| 1 | prep  v [ing] |  |  |  |  | ▓ | ▓ |  |  |
| 5 | art [def]  n [pl ,f] | ▓ | ▓ |  |  |  |  |  |  |
| 8 | adj |  |  |  | ▓ |  |  |  |  |
| 10 | poss [pl,m,1] n [pl |  |  |  |  |  |  | ▓ | ▓ |
| 15 | dummy |  |  | ▓ |  |  |  |  |  |

*5.5 Selecting the Optimum Path*

Fortunately, in this example, only one possible path of chunk templates can cover the whole word patterns of the given sentence. This happened due to the positive impact of the pruning process applied to the correspondence matrix. This final path will be considered as the best template path that represents the given sentence.

| Best Template Path |
|---|
| Chunk5 + chunk15 + chunk8 + chunk1 + chunk10 |

*5.6 Extracting the corresponding Arabic templates*

| No | English Chunk Template Path | Arabic Chunk Templates |
|---|---|---|
| 1 | prep  v [ing] | (prep1)  (v1 [source]) |
| 5 | art [def]  n [pl ,f] | (add [ال]  n1 [pmean]) |
| 8 | adj | (adj1 [s ,f]) |
| 10 | poss [pl,m,1] n [pl ,f] | (n1 [pmean]  add [نا]) |
| 15 | dummy | dummy |

*5.7 Generating the Target Sentence*

| No | Arabic template | Arabic chunk |
|---|---|---|
| 1 | (prep1)  (v1 [source]) | لبناء |
| 5 | (add [ال]  n1 [pmean]) | البروتينيات |
| 8 | (adj1 [s ,f]) | ضرورية |
| 10 | (n1 [pmean]  add [نا]) | أجسامنا |
| 15 | dummy |  |

The final translation is:

" البروتينيات ضرورية لبناء أجسامنا "

## 6. Example 2
Another example is proposed to illustrate how correspondence matrix is pruned and the role of the directed graph in selecting the optimum chunk path.

Assume the English sentence is "the good diet depends on the balance between the main elements for our bodies".

The correspondence matrix is given by:

| No | w1 | w2 | w3 | w4 | w5 | w6 | w7 | w8 | w9 | w10 | w11 | w12 | w13 | w14 |
|---|---|---|---|---|---|---|---|---|---|---|---|---|---|---|
|  | the | good | diet | depends | on | the | balance | between | the | main | elements | for | our | bodies |
|  | Art | Adj | N | V | Prep | Def | N | Prep | Art | Adj | N | Prep | Poss | N |
| 1 | ▓ | ▓ | | | | | | | | | | | | |
| 2 | ▓ | ▓ | ▓ | | | | | | | | | | | |
| 3 | ▓ | ▓ | ▓ | ▓ | ▓ | ▓ | | | | | | | | |
| 4 | | ▓ | | | | | | | | | | | | |
| 5 | | ▓ | ▓ | | | | | | | | | | | |
| 6 | | | ▓ | | | | | | | | | | | |
| 7 | | | ▓ | ▓ | ▓ | | | | | | | | | |
| 8 | | | | ▓ | ▓ | | | | | | | | | |
| 9 | | | | | ▓ | | | | | | | | | |
| 10 | | | | | ▓ | ▓ | ▓ | | | | | | | |
| 11 | | | | | | ▓ | | | | | | | | |
| 12 | | | | | | | ▓ | ▓ | ▓ | | | | | |
| 13 | | | | | | | ▓ | | | | | | | |
| 14 | | | | | | | | ▓ | | | | | | |
| 15 | | | | | | | | | ▓ | ▓ | ▓ | | | |
| 16 | | | | | | | | | ▓ | ▓ | ▓ | ▓ | | |
| 17 | | | | | | | | | | ▓ | ▓ | | | |
| 18 | | | | | | | | | | | ▓ | ▓ | | |
| 19 | | | | | | | | | | | | ▓ | | |
| 20 | | | | | | | | | | | | ▓ | ▓ | ▓ |
| 21 | | | | | | | | | | | | | ▓ | ▓ |
| 22 | | | | | | | | | | | | | ▓ | ▓ |
| 23 | | | | | | | | | | | | | ▓ | ▓ |
| 24 | ▓ | ▓ | ▓ | | | | | | | | | | | |
| 25 | | | | ▓ | ▓ | | | | | | | | | |
| 26 | | | | | | ▓ | | | | | | | | |
| 27 | | | | | | | | ▓ | ▓ | ▓ | ▓ | | | |

| | Repeated chunks |
|---|---|
| | Unreachable |
| | Dead End |

After pruning the matrix (removing repeated, unreachable and dead end chunks), the matrix is reduced to the following:

| No | w1 the | w2 good | w3 diet | w4 depends | w5 on | w6 the | w7 balance | w8 between | w9 the | w10 main | w11 elements | w12 for | w13 our | w14 bodies |
|---|---|---|---|---|---|---|---|---|---|---|---|---|---|---|
| 1 | ▓ | ▓ | | | | | | | | | | | | |
| 2 | ▓ | ▓ | ▓ | | | | | | | | | | | |
| 3 | ▒ | ▒ | ▒ | ▒ | ▒ | ▒ | | | | | | | | |
| 6 | | | ▓ | | | | | | | | | | | |
| 7 | | | ▓ | ▓ | ▓ | | | | | | | | | |
| 8 | | | | ▓ | ▓ | | | | | | | | | |
| 11 | | | | | | ▓ | ▓ | | | | | | | |
| 12 | | | | | | ▓ | ▓ | ▓ | | | | | | |
| 13 | | | | | | | ▒ | ▒ | | | | | | |
| 14 | | | | | | | | ▓ | | | | | | |
| 15 | | | | | | | | | ▓ | ▓ | ▓ | | | |
| 16 | | | | | | | | | ▓ | ▓ | ▓ | ▓ | | |
| 19 | | | | | | | | | | | | ▓ | | |
| 20 | | | | | | | | | | | | ▒ | ▒ | ▒ |
| 21 | | | | | | | | | | | | | ▓ | ▓ |
| 27 | | | | | | | | ▒ | ▒ | ▒ | ▒ | | | |

**The best path chunks**

The directed graph representation can be summarized as::

| Directed Graph Paths |
|---|
| chunk1 + chunk6 + chunk8 + chunk11 + chunk14 + chunk15 + chunk19 + chunk21 |
| chunk2 + chunk8 + chunk11 + chunk14 + chunk15 + chunk19 + chunk21 |
| chunk3 + chunk13 + chunk14 + chunk15 + chunk19 + chunk21 |
| chunk1 + chunk7 + chunk11 + chunk14 + chunk15 + chunk19 + chunk21 |
| chunk1 + chunk6 + chunk8 + chunk12 + chunk15 + chunk19 + chunk21 |
| chunk1 + chunk6 + chunk8 + chunk11 + chunk27 + chunk19 + chunk21 |
| chunk1 + chunk6 + chunk8 + chunk11 + chunk14 + chunk16 + chunk21 |
| chunk1 + chunk6 + chunk8 + chunk11 + chunk14 + chunk15 + chunk20 |
| chunk2 + chunk8 + chunk12 + chunk15 + chunk19 + chunk21 |
| chunk2 + chunk8 + chunk11 + chunk27 + chunk19 + chunk21 |
| chunk2 + chunk8 + chunk11 + chunk14 + chunk16 + chunk21 |
| chunk2 + chunk8 + chunk11 + chunk14 + chunk15 + chunk20 |
| chunk3 + chunk13 + chunk27 + chunk19 + chunk21 |
| chunk3 + chunk13 + chunk14 + chunk16 + chunk21 |
| chunk3 + chunk13 + chunk14 + chunk15 + chunk20 |
| chunk1 + chunk7 + chunk12 + chunk15 + chunk19 + chunk21 |
| chunk1 + chunk7 + chunk11 + chunk27 + chunk19 + chunk21 |
| chunk1 + chunk7 + chunk11 + chunk14 + chunk16 + chunk21 |
| chunk1 + chunk7 + chunk11 + chunk14 + chunk15 + chunk20 |
| chunk1 + chunk6 + chunk8 + chunk12 + chunk16 + chunk21 |
| chunk1 + chunk6 + chunk8 + chunk12 + chunk15 + chunk20 |
| chunk1 + chunk6 + chunk8 + chunk11 + chunk27 + chunk20 |
| chunk2 + chunk8 + chunk12 + chunk16 + chunk21 |
| chunk2 + chunk8 + chunk12 + chunk15 + chunk20 |
| chunk2 + chunk8 + chunk11 + chunk27 + chunk20 |
| **chunk3 + chunk13 + chunk27 + chunk20** |
| chunk1 + chunk7 + chunk12 + chunk16 + chunk21 |
| chunk1 + chunk7 + chunk12 + chunk15 + chunk20 |
| chunk1 + chunk7 + chunk11 + chunk27 + chunk20 |

**Searching for the optimum path:**

Applying the heuristic that "*fewest path's chunks and longer chunks are preferred leads to the selection of the path*"

**chunk3 + chunk13 + chunk27 + chunk20**

**Conclusion**

In this paper, we have presented a best template match technique for Example Based Machine Translation. The technique relies on using a tagged parallel corpus, aligned at chunk level. This allows much more flexibility in using the same chunk in many different sentences, and hence reduces the required corpus size. The technique tries all possible combinations of input words, and gets its corresponding chunks. A correspondence matrix is built, which maps the coverage of the found chunks to input English sentence. The matrix is then tuned to get the candidate chunks. The proposed technique then builds a directed graph that represents all suitable chunks which covers the whole input sentence (paths). Optimum path is selected to be a path with a minimum number of chunks, and avoids dummy chunks.

In this research, the translation is done at a smaller granularity chunk level, which allows the reduction of the required parallel corpus. More reduction in corpus size is also achieved by searching on word's features rather than the word itself. Using tagged corpus is most suitable for Arabic language, which takes into account the different morphological forms, which enhances the quality of translation.